\documentclass[11pt]{article}

\usepackage[final]{acl}

\usepackage{times}
\usepackage{latexsym}

\usepackage[T1]{fontenc}
\usepackage[utf8]{inputenc}
\usepackage{microtype}
\usepackage{inconsolata}
\usepackage{graphicx}

\usepackage{amssymb}
\usepackage{amsmath}
\usepackage{booktabs}
\usepackage{multirow}
\usepackage[table]{xcolor}
\usepackage{makecell}
\usepackage{listings}
\usepackage{pifont}

\newcommand{\samethanks}[1][\value{footnote}]{\footnotemark[#1]}

\title{Controllable Affective Generation via Latent Vector Steering}

\author{Xixian Yong\textsuperscript{\rm 1} \quad Siyuan Chang\textsuperscript{\rm 1} \quad Yingying Zhang\textsuperscript{\rm 4} \quad Xian Wu\textsuperscript{\rm 4}\thanks{Corresponding authors: Xiao Zhou and Xian Wu.} \quad Xiao Zhou\textsuperscript{\rm 1,2,3}\samethanks\\
\textsuperscript{\rm 1}Gaoling School of Artificial Intelligence, Renmin University of China \\
\textsuperscript{\rm 2}Beijing Key Laboratory of Research on Large Models and Intelligent Governance \\
\textsuperscript{\rm 3}Engineering Research Center of Next-Generation Intelligent Search and Recommendation, MOE \\
\textsuperscript{\rm 4}Tencent Jarvis Lab \\
\texttt{\{xixianyong,xiaozhou\}@ruc.edu.cn} \quad 
\texttt{kevinxwu@tencent.com}
}

\begin{document}

\maketitle

\begin{abstract}
Large Language Models (LLMs) often produce emotionally flattened responses after alignment, limiting their effectiveness in affect-sensitive applications. In this paper, we propose EmoVec, a lightweight framework for controllable affective generation via latent vector steering. EmoVec extracts emotion-specific directions from paired neutral and emotion-conditioned responses using contrastive activation addition, and further refines them through task-specific debiasing and principal subspace removal. During inference, these vectors are injected into the final residual stream with static or scenario-adaptive scaling, enabling continuous control over emotional intensity without updating model weights. Experiments across three LLMs and eight emotions show that EmoVec consistently improves emotional salience while largely preserving semantic content, fluency, and coherence. Ablation studies and human evaluation further confirm the effectiveness of vector purification and adaptive scaling, establishing EmoVec as a practical inference-time method for affective control in deployed LLMs. Code and data are available at \url{https://github.com/chicosirius/EmoVec}.
\end{abstract}

\section{Introduction}
The advent of Large Language Models (LLMs) has revolutionized Natural Language Processing (NLP)~\citep{brown2020language}, enabling systems that exhibit remarkable proficiency in reasoning, coding, and general knowledge retrieval. Despite these cognitive leaps, a significant gap remains in the domain of emotional intelligence~\citep{sabour2024emobench}. While current models can simulate emotions when explicitly prompted, their default outputs, which are heavily conditioned by Reinforcement Learning from Human Feedback (RLHF), often suffer from emotional flattening~\citep{Kirk2023UnderstandingTE,ibrahim2025training}. In pursuit of safety and harmlessness, RLHF tends to compress the distribution of model outputs towards neutrality, often manifesting as responses that rely on generic reassurance phrases, excessive hedging, or well-documented sycophantic behaviors that prioritize agreement over context-sensitive expression~\citep{dahlgren2025helpful,gonzalez2025reinforcement}. This alignment tax~\citep{askell2021general,lin2024mitigating} limits the applicability of LLMs in fields requiring high affective nuance, such as mental health support, creative writing, and empathetic human-computer interaction~\citep{DBLP:conf/nips/YongZZLZW25,DBLP:conf/www/YongSWZ26,guo2026counterfactual,zhang2026humanvaluesmatterinvestigating}.

\begin{figure}[t]
    \centering
    \includegraphics[width=1\linewidth]{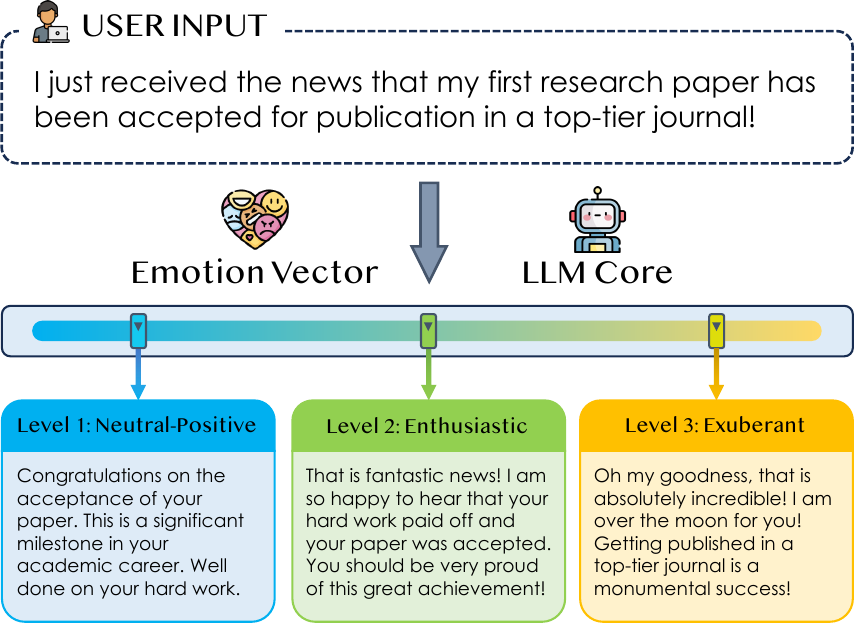}
    \caption{\textbf{Conceptual illustration of controllable affective generation.} Through the proposed latent vector steering mechanism, the model's output is modulated along a \textit{Happiness Gradient}. This results in three distinct responses that maintain semantic consistency with the input while exhibiting progressively higher levels of emotional intensity, ranging from professional acknowledgment (Level 1) to exuberant celebration (Level 3).}
    \label{fig:example}
\end{figure}

Current approaches to mitigating this limitation primarily rely on prompt engineering or supervised fine-tuning (SFT). Prompt-based strategies~\citep{li2023large} (e.g., "Act as an empathetic therapist") are notoriously brittle, consuming valuable context window space and yielding inconsistent results sensitive to lexical variation~\citep{wang2022interpretability,miehling2025evaluating}. Supervised fine-tuning, while effective, requires large-scale labeled datasets and substantial computational resources, and it may introduce risks such as catastrophic forgetting or degradation of the model’s general capabilities~\citep{lin2024mitigating,zhu2026mohobench}. These limitations highlight the need for a lightweight and controllable method that enables affective generation without retraining the model.

To ground our approach, we first investigate where and how emotion is represented within LLMs. We conduct a probing analysis by feeding texts with varying emotional polarities into the model and training linear classifiers on the hidden states of each layer. Our results reveal a consistent and interpretable pattern: \textbf{representations of emotional states become increasingly linearly separable in the middle-to-late layers of the model.} This finding aligns with and extends recent concurrent work. For instance, \citet{cintas2025localizing} show that persona-specific representations are most separable in the final third of model layers, while \citet{ju2025probing} demonstrate that personality traits emerge progressively and crystallize in upper layers. Together, these results suggest that while early layers encode syntax and shallow semantics, higher layers capture abstract affective and persona-related attributes~\citep{rogers2020primer}. This layer-wise structure provides a precise and principled intervention point for affective control~\citep{guo2026not}.

Motivated by this observation, we propose EmoVec, a framework for controllable affective generation via latent vector manipulation. Building on Representation Engineering (RepE)~\citep{zou2023representation}, which represents high-level semantics as linear directions in activation space~\citep{park2023linear,turner2023steering}, EmoVec introduces a principled approach to isolate and purify emotion-specific vectors while disentangling them from task semantics. Injected at inference with adjustable intensity, these vectors enable fine-grained control without weight updates or prompt engineering.

EmoVec extracts emotional steering vectors using Contrastive Activation Addition (CAA)~\citep{rimsky2024steering,chen2025persona}. We construct paired prompt–response trajectories matched in semantic content but differing in emotional affect, and compute differences in their mean activations at targeted layers, yielding vectors that capture affective variation while minimizing task-related confounds. These vectors are applied during inference with a tunable scaling factor to control both emotion type and intensity.

As shown in Figure~\ref{fig:example}, this mechanism enables continuous modulation of affective intensity for a fixed input while preserving semantic consistency. Such fine-grained controllability allows LLMs to adapt their emotional expression to diverse contextual demands, including professional neutrality, empathetic support, and expressive creativity. This capability is particularly valuable for human-facing applications such as psychological support, creative writing, and personalized conversational agents~\citep{zhou2025tricolore,guo2025sorex}.

Our contributions are as follows:
\begin{itemize}
    \item We provide empirical evidence validating the layer-wise emergence of emotional representations in LLMs, reinforcing the Linear Representation Hypothesis in the context of affective computing.
    \item We develop a robust pipeline for extracting and verifying emotional steering vectors using contrastive examples.
    \item We implement a mechanism for dynamic, fine grained control over emotional intensity, allowing models to adapt their affective expressiveness to scenario specific demands.
\end{itemize}

\begin{figure*}[t]
    \centering
    \includegraphics[width=\linewidth]{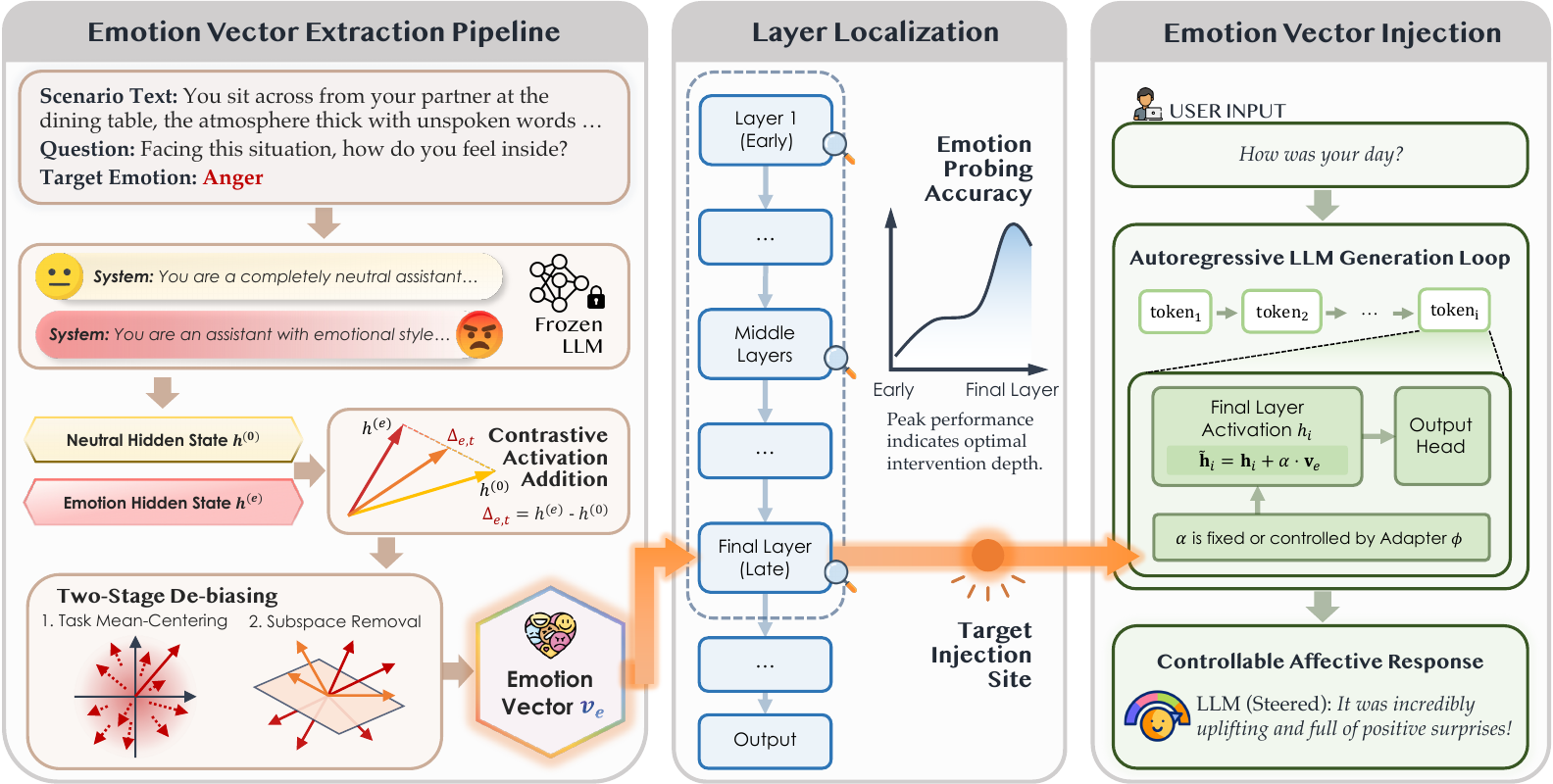}
    \caption{\textbf{Overview of the latent vector steering framework.} The pipeline consists of (1) Emotion Vector Extraction using CAA and two stage debiasing to isolate purified signals , (2) Layer Localization via linear probing to identify the optimal intervention site at the final layer , and (3) Emotion Vector Injection into the final residual stream where intensity is dynamically modulated by a scenario adaptive adapter.}
    \label{fig:framework}
\end{figure*}

\section{Related Work}
\paragraph{Affective Computing in Language Models.}
Affective computing aims to enable machines to recognize and generate human emotional states. With the advent of LLMs, research has shifted toward assessing their emergent affective capabilities. Studies indicate that models like the GPT-series can estimate valence, arousal, and perform appraisal-based emotion elicitation purely from linguistic data~\citep{broekens2023fine,zhang2024affective}. LLMs also optimize emotion annotation workflows by assisting humans in identifying low-quality labels, thereby enhancing downstream performance~\citep{niu2025rethinking}. While these models excel at capturing general emotional polarity and dialogue-based recognition, they still struggle with fine-grained distinctions and multimodal contexts~\citep{sabour2024emobench,sorin2024large,castro2025emotion}. Nevertheless, LLMs show promise in generating synthetic emotional datasets and performing socio-emotional tasks like empathy evaluation~\citep{kaplan2025can,dong2025integrating}.

\paragraph{Representation Engineering and Activation Steering.}
Representation engineering manipulates hidden representations to control LLM behavior without updating model parameters \citep{zou2023representation,turner2023steering}. Prior work shows that latent directions can steer properties such as writing style and sentiment \citep{diallo2025style,farooq2025sentiment}. Other studies extract steering vectors directly from pretrained language models \citep{subramani2022extracting} or identify persona-related representations in activation space \citep{chen2025persona}. Recent studies further suggest that emotion- and persona-related information becomes increasingly separable in middle-to-late layers of LLMs \citep{cintas2025localizing,ju2025probing}. Emotion-specific neurons and affective representations have also been investigated from a mechanistic perspective \citep{emotionneurons2025,emotioninference2025,DBLP:conf/acl/YongLY0025}. 

Different from prior work, EmoVec focuses on fine-grained emotion-intensity control rather than coarse sentiment or generic style transfer. We further introduce task-specific debiasing to reduce semantic contamination and evaluate semantic preservation under different steering strengths.

\section{Emotion Vector Extraction}
\label{sec:extraction}
We formulate controllable affective generation as emotion-intensity modulation under semantic preservation constraints. Our goal is to increase target emotional intensity while preserving the original semantic intent. We extract emotion-specific latent directions from paired neutral and emotion-conditioned responses and reduce task-specific semantic variation before inference-time steering.

For each target emotion $e$, we aim to obtain a vector $v_e \in \mathbb{R}^d$ that captures the representation shift induced by expressing emotion $e$ while remaining robust to scenario-specific semantics.

\subsection{Notation and Preliminaries}
We assume a dataset of \(N\) scenario tasks. For each task \(t\) and each target emotion class \(e \in \mathcal{E}\), the LLM is prompted twice for the same scenario: \textbf{1) neutral response}, from which we extract a representation vector \(\mathbf{h}^{(0)}_{e,t}\in\mathbb{R}^d\); \textbf{2) emotion-conditioned response}, yielding \(\mathbf{h}^{(e)}_{e,t}\in\mathbb{R}^d\).

In practice the representation \(\mathbf{h}\) is computed by averaging token-level hidden activations over the response tokens. For brevity we denote the pair for task \(t\) and emotion \(e\) simply as \((\mathbf{h}^{(0)}_{e,t}, \mathbf{h}^{(e)}_{e,t})\). We further assume that each generated response is associated with scalar quality scores \(s^{(0)}_{e,t}\) (neutral) and \(s^{(e)}_{e,t}\) (emotional) produced by a LLM-based judge. To ensure the robustness of the evaluation, we conduct a validation study checking the agreement between the model's scores and human judgments (see Appendix~\ref{app:consistency} for details).

\subsection{Scenario Construction}
To construct paired affective scenarios while controlling semantic content, we sample short social and commonsense \emph{seeds} from Social Chemistry \citep{forbes2020social}, NormBank \citep{ziems2023normbank}, and Social IQa \citep{sap2019socialiqa}. These seeds cover four domains: Work \& Productivity, Intimate Relationships, Public \& Societal Interactions, and Personal Feelings.

For each seed, we use an LLM-assisted rewriting pipeline to generate a complete scenario, a corresponding question, and a target emotion. We require the scenario to support the target affect without explicitly naming it and to remain compatible with both neutral and emotion-conditioned responses. The generated scenarios are manually filtered for semantic clarity, emotional plausibility, and absence of emotion leakage. We retain 160 scenarios for each emotion (1,280 total), split evenly between vector extraction and evaluation. The same scenario set is reused across all evaluated LLMs, while emotion vectors are extracted separately for each model.

\subsection{Contrastive Activation Addition}
To reduce noise caused by poor or malformed generations, we keep only high-quality pairs. Formally, let \(\tau_0,\tau_e\) be thresholds for neutral and emotional quality respectively. We retain the index set:
\begin{equation}
    \mathcal{I} \;=\; \{(e,t)\ :\ s^{(0)}_{e,t} \ge \tau_0 \ \wedge\ s^{(e)}_{e,t} \ge \tau_e\}.
\end{equation}
In our experiments we set each threshold to a chosen percentile of the corresponding score distribution. For each retained pair \((e,t)\in\mathcal{I}\) we define the \emph{emotion shift vector}:
\begin{equation}
\Delta_{e,t} \;=\; \mathbf{h}^{(e)}_{e,t} - \mathbf{h}^{(0)}_{e,t} \in \mathbb{R}^d.
\label{eq:delta}
\end{equation}
This vector captures how the model's internal representation moves when producing an emotion-conditioned response instead of a neutral response for the same scenario.

\subsection{Task-Specific Debiasing}
\label{sec:debiasing}
While the contrastive shifts $\Delta_{e,t}$ (Eq. \ref{eq:delta}) isolate the representation change between emotional and neutral states, they may still be contaminated by task-specific semantics, such as interpersonal dynamics, narrative styles, or topical domains. To extract a purified emotion signal, we employ a two-stage debiasing procedure consisting of mean-centering and subspace removal.

\paragraph{First-order Task Centering.} 
We first mitigate first-order task bias by computing a per-emotion average across scenarios. For each emotion $e$, the task mean shift is defined as:
\begin{equation}
\bar{\Delta}_{e} \;=\; \frac{1}{|\mathcal{T}_e|}\sum_{t\in\mathcal{T}_e} \Delta_{e,t},
\end{equation}
where $\mathcal{T}_e$ represents all tasks related to emotion $e$. The task-centered shift is then obtained:
\begin{equation}
\Delta'_{e,t} \;=\; \Delta_{e,t} - \bar{\Delta}_{e}.
\label{eq:task_center}
\end{equation}
Intuitively, $\bar{\Delta}_e$ represents the centroid of the representational shift for emotion $e$ across its task distribution. By subtracting this mean, we obtain $\Delta'_{e,t}$ to isolate the intraclass variance. This term represents the noise induced by scenario specific semantics, such as topical or stylistic variations, relative to the core direction of the emotion.

\paragraph{Subspace Removal via Orthogonal Projection.} 
Even after centering, task-specific semantic variations may still dominate the variance in a low-dimensional subspace. To further suppress such variation, we excise the task-dominated subspace using Principal Component Analysis (PCA).

Let $\mathbf{D} \in \mathbb{R}^{M \times d}$ be the matrix formed by stacking all centered shift vectors $\Delta'_{e,t}$ as rows, where $M = |\mathcal{I}|$ is the total number of retained pairs. We perform PCA on $\mathbf{D}$ to identify the top-$k$ principal components:
\begin{equation}
\mathbf{U}_k = [\mathbf{u}_1, \mathbf{u}_2, \dots, \mathbf{u}_k] \in \mathbb{R}^{d \times k},
\end{equation}
where each $\mathbf{u}_i$ represents a primary direction of task-related semantic variance. We then project the centered shifts onto the orthogonal complement of the subspace spanned by $\mathbf{U}_k$:
\begin{equation}
\hat{\Delta}_{e,t} \;=\; \Delta'_{e,t} - \mathbf{U}_k \mathbf{U}_k^\top \Delta'_{e,t}.
\label{eq:residual}
\end{equation}

The resulting residual vector $\hat{\Delta}_{e,t}$ is orthogonal to the dominant task-related directions, effectively concentrating the emotion-related variation.

\subsection{Principal Direction Aggregation}
Given residual vectors \(\{\hat{\Delta}_{e,t}\}\), we obtain an estimated per-emotion direction \(\mathbf{v}_e\) by aggregating across tasks \(t\) that share emotion \(e\). Specifically, we use Principal direction aggregation, which concatenates residuals for emotion \(e\) into a matrix \(R_e\) and compute the top principal component:
  \begin{equation}
  \mathbf{v}_e \;=\; \arg\max_{\|\mathbf{w}\|_2=1} \mathbf{w}^\top \, \mathrm{Cov}(R_e) \, \mathbf{w},
  \label{eq:per_emotion_pca}
  \end{equation}
i.e., choose \(\mathbf{v}_e\) as the first eigenvector of the residual covariance for emotion \(e\).

\section{Emotion Vectors Intervention}
\subsection{Layer Localization via Linear Probing}
\label{sec:layer_localization}
\begin{figure}[ht]
    \centering
    \includegraphics[width=\linewidth]{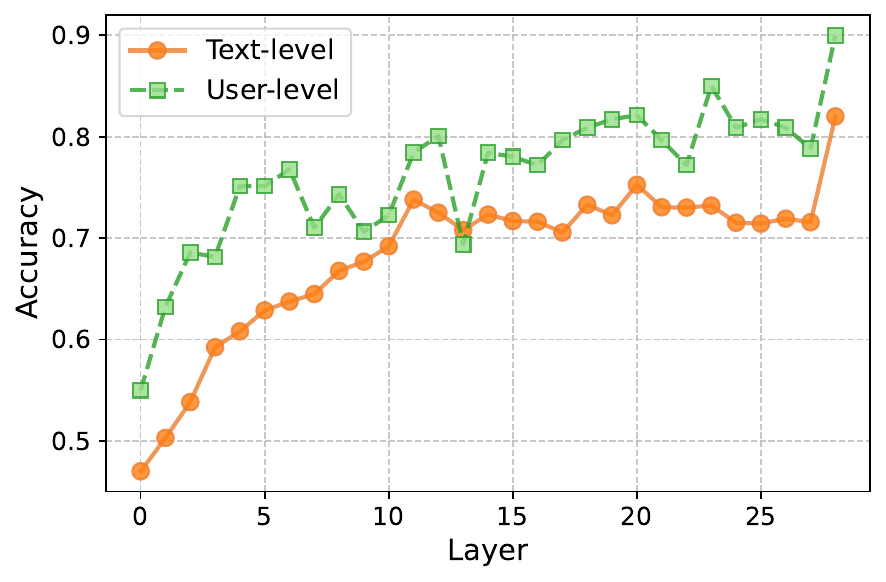}
    \caption{\textbf{Prediction accuracy across layers.} Text-level probing treats each response as an independent sample, while user-level probing averages representations over responses associated with the same scenario or user context before classification.}
    \label{fig:layer_analysis}
\end{figure}

Although the extraction procedure can yield a candidate emotion vector $\mathbf{v}_e^{(l)}$ for every layer $l \in \{1, \dots, L\}$, our preliminary experiments (Figure~\ref{fig:layer_analysis}) indicate that emotional representations are not uniformly distributed throughout the model layers.

For each layer $l$, we trained a logistic regression classifier $\mathcal{C}_l$ to predict the emotion category $e$ based on the centered hidden states. While emotional features begin to emerge in the middle layers, we observe that the modeling of emotional states reaches its peak crystallization in the final layer of the model. This layer serves as the ultimate semantic bottleneck where abstract emotional concepts are most linearly separable and directly influence the output logits. Consequently, we concentrate our intervention efforts exclusively on the final layer $L$ to maximize steering efficacy while minimizing cumulative noise across the residual stream.

\subsection{Latent Vector Steering}
\label{sec:steering_mechanism}
During the inference phase, we steer the model by injecting the purified emotion vector $\mathbf{v}_e$ directly into the final residual stream. Unlike prompt engineering which attempts to influence the model through input tokens, our method performs a direct intervention on the internal activation $\mathbf{h}_{i}$ at each token step $i$.

Formally, let $\mathbf{h}_{i}$ denote the original activation of the final layer given the current context. The steered activation $\tilde{\mathbf{h}}_{i}$ is computed as follows:
\begin{equation}
\tilde{\mathbf{h}}_{i} \;=\; \mathbf{h}_{i} + \alpha \cdot \mathbf{v}_e,
\label{eq:steering}
\end{equation}
In this equation, $\alpha \in \mathbb{R}^+$ represents a scalar steering coefficient that modulates the intensity of the emotional infusion. This intervention is applied during every forward pass of the autoregressive generation process, which effectively biases the output probability distribution towards tokens that semantically align with the target emotion.

\subsection{Scenario-Adaptive Intensity Control}
\label{sec:adaptive_steering}
Static steering coefficients often fail to accommodate the diverse emotional demands of different contexts. To achieve precise control, we introduce a learnable adapter $\phi$ designed to modulate intervention intensity based on scenario requirements.

This lightweight adapter is trained to map the scenario context to an optimal scaling factor. Formally, for a given scenario context $\mathbf{c}$, the adapter $\phi$ generates a scenario specific coefficient $\lambda_{\mathbf{c}} = \phi(\mathbf{c})$. The steering operation at token step $i$ is then defined as:
\begin{equation}
\tilde{\mathbf{h}}_{i} \;=\; \mathbf{h}_{i} + (\lambda_{\mathbf{c}} \cdot \|\mathbf{h}_{i}\|_2) \cdot \mathbf{v}_e,
\end{equation}
where the intervention strength is jointly determined by the learned scenario importance and the instantaneous activation norm. 

This framework allows the model to intelligently allocate emotional strength according to contextual sensitivity. By optimizing the adapter, the system maintains high affective expressiveness in pertinent scenarios while preserving semantic neutrality in objective contexts, thereby ensuring linguistic integrity and preventing semantic collapse.

\begin{table*}[t]
  \centering
  \caption{\textbf{Evaluation of steering controllability across different emotions and model scales.} The table displays absolute scores and relative gains ($\eta$) for three base models under varying steering magnitudes ($\alpha$). The results demonstrate a consistent positive correlation between the steering coefficient and the resulting emotional salience.}
  \resizebox{\linewidth}{!}{
    \begin{tabular}{clrrrrrrrr>{\columncolor[rgb]{0.96,0.96,0.96}}r}
    \toprule
    \multicolumn{2}{c}{\textbf{Emotion}} & \multicolumn{1}{c}{\textbf{Anger}} & \multicolumn{1}{c}{\textbf{Anticipation}} & \multicolumn{1}{c}{\textbf{Disgust}} & \multicolumn{1}{c}{\textbf{Fear}} & \multicolumn{1}{c}{\textbf{Joy}} & \multicolumn{1}{c}{\textbf{Sadness}} & \multicolumn{1}{c}{\textbf{Surprise}} & \multicolumn{1}{c}{\textbf{Trust}} & \multicolumn{1}{c}{\textbf{Avg.}} \\
    \midrule
    \rowcolor[rgb]{ .929,  .929,  .929} \multicolumn{11}{c}{\textbf{Qwen2.5-7B-Instruct}} \\
    \multicolumn{2}{c}{w/o injection } & 65.14  & 70.50  & 54.69  & 77.80  & 80.25  & 60.30  & 72.10  & 75.55  & 69.54  \\
    \multirow{2}[0]{*}{$\alpha=5$} & score & 67.91  & 71.92  & 55.26  & 83.40  & 84.00  & 63.15  & 75.95  & 78.40  & \textbf{72.50}  \\
          & $\eta$ & {\color{gray}\textit{+4.25\%}} & {\color{gray}\textit{+2.01\%}} & {\color{gray}\textit{+1.04\%}} & {\color{gray}\textit{+7.20\%}} & {\color{gray}\textit{+4.67\%}} & {\color{gray}\textit{+4.73\%}} & {\color{gray}\textit{+5.34\%}} & {\color{gray}\textit{+3.77\%}} & {\color{gray}\textit{+4.26\%}} \\
    \multirow{2}[0]{*}{$\alpha=10$} & score & 70.18  & 83.42  & 55.52  & 83.38  & 90.50  & 68.20  & 77.20  & 81.10  & \textbf{76.19}  \\
          & $\eta$ & {\color{gray}\textit{+7.74\%}} & {\color{gray}\textit{+18.33\%}} & {\color{gray}\textit{+1.52\%}} & {\color{gray}\textit{+7.17\%}} & {\color{gray}\textit{+12.77\%}} & {\color{gray}\textit{+13.10\%}} & {\color{gray}\textit{+7.07\%}} & {\color{gray}\textit{+7.35\%}} & {\color{gray}\textit{+9.56\%}} \\
    \multirow{2}[1]{*}{$\alpha=50$} & score & 83.44  & 83.57  & 77.75  & 86.37  & 92.15  & 75.40  & 88.90  & 85.95  & \textbf{84.19}  \\
          & $\eta$ & {\color{gray}\textit{+28.09\%}} & {\color{gray}\textit{+18.54\%}} & {\color{gray}\textit{+42.16\%}} & {\color{gray}\textit{+11.02\%}} & {\color{gray}\textit{+14.83\%}} & {\color{gray}\textit{+25.04\%}} & {\color{gray}\textit{+23.30\%}} & {\color{gray}\textit{+13.77\%}} & {\color{gray}\textit{+21.07\%}} \\
    \midrule
    \rowcolor[rgb]{ .929,  .929,  .929} \multicolumn{11}{c}{\textbf{Llama3.1-8B-Instruct}} \\
    \multicolumn{2}{c}{w/o injection} & 68.45  & 77.91  & 51.33  & 79.28  & 81.17  & 56.29  & 74.30  & 75.22  & 70.49  \\
    \multirow{2}[0]{*}{$\alpha=5$} & score & 72.88  & 78.49  & 61.60  & 81.95  & 88.11  & 61.75  & 74.62  & 83.65  & \textbf{75.38}  \\
          & $\eta$ & {\color{gray}\textit{+6.47\%}} & {\color{gray}\textit{+0.74\%}} & {\color{gray}\textit{+20.01\%}} & {\color{gray}\textit{+3.37\%}} & {\color{gray}\textit{+8.55\%}} & {\color{gray}\textit{+9.70\%}} & {\color{gray}\textit{+0.43\%}} & {\color{gray}\textit{+11.21\%}} & {\color{gray}\textit{+6.94\%}} \\
    \multirow{2}[0]{*}{$\alpha=10$} & score & 76.12  & 82.95  & 54.21  & 88.89  & 91.54  & 72.33  & 82.15  & 87.18  & \textbf{79.42}  \\
          & $\eta$ & {\color{gray}\textit{+11.21\%}} & {\color{gray}\textit{+6.47\%}} & {\color{gray}\textit{+5.61\%}} & {\color{gray}\textit{+12.12\%}} & {\color{gray}\textit{+12.78\%}} & {\color{gray}\textit{+28.50\%}} & {\color{gray}\textit{+10.57\%}} & {\color{gray}\textit{+15.90\%}} & {\color{gray}\textit{+12.67\%}} \\
    \multirow{2}[1]{*}{$\alpha=50$} & score & 80.55  & 85.11  & 63.08  & 90.42  & 93.20  & 74.58  & 84.44  & 89.15  & \textbf{82.57}  \\
          & $\eta$ & {\color{gray}\textit{+17.68\%}} & {\color{gray}\textit{+9.24\%}} & {\color{gray}\textit{+22.89\%}} & {\color{gray}\textit{+14.05\%}} & {\color{gray}\textit{+14.82\%}} & {\color{gray}\textit{+32.49\%}} & {\color{gray}\textit{+13.65\%}} & {\color{gray}\textit{+18.52\%}} & {\color{gray}\textit{+17.14\%}} \\
    \midrule
    \rowcolor[rgb]{ .929,  .929,  .929} \multicolumn{11}{c}{\textbf{Qwen2.5-70B-Instruct}} \\
    \multicolumn{2}{c}{w/o injection} & 67.04  & 71.98  & 56.11  & 78.55  & 81.63  & 61.25  & 77.10  & 76.95  & 71.33  \\
    \multirow{2}[0]{*}{$\alpha=5$} & score & 69.81  & 73.15  & 60.15  & 83.08  & 85.05  & 63.85  & 79.55  & 77.10  & \textbf{73.97}  \\
          & $\eta$ & {\color{gray}\textit{+4.13\%}} & {\color{gray}\textit{+1.63\%}} & {\color{gray}\textit{+7.20\%}} & {\color{gray}\textit{+5.77\%}} & {\color{gray}\textit{+4.19\%}} & {\color{gray}\textit{+4.24\%}} & {\color{gray}\textit{+3.18\%}} & {\color{gray}\textit{+0.19\%}} & {\color{gray}\textit{+3.70\%}} \\
    \multirow{2}[0]{*}{$\alpha=10$} & score & 73.08  & 82.55  & 62.05  & 83.15  & 92.11  & 69.10  & 80.01  & 78.05  & \textbf{77.51}  \\
          & $\eta$ & {\color{gray}\textit{+9.01\%}} & {\color{gray}\textit{+14.68\%}} & {\color{gray}\textit{+10.59\%}} & {\color{gray}\textit{+5.86\%}} & {\color{gray}\textit{+12.84\%}} & {\color{gray}\textit{+12.82\%}} & {\color{gray}\textit{+3.77\%}} & {\color{gray}\textit{+1.43\%}} & {\color{gray}\textit{+8.66\%}} \\
    \multirow{2}[1]{*}{$\alpha=50$} & score & 85.95  & 86.81  & 77.58  & 87.89  & 93.58  & 91.15  & 85.99  & 76.85  & \textbf{85.73}  \\
          & $\eta$ & {\color{gray}\textit{+28.21\%}} & {\color{gray}\textit{+20.60\%}} & {\color{gray}\textit{+38.26\%}} & {\color{gray}\textit{+11.89\%}} & {\color{gray}\textit{+14.64\%}} & {\color{gray}\textit{+48.82\%}} & {\color{gray}\textit{+11.53\%}} & {\color{gray}\textit{-0.13\%}} & {\color{gray}\textit{+20.19\%}} \\
    \bottomrule
    \end{tabular}}
  \label{tab:addlabel}%
\end{table*}%

\section{Experiments}
In this section, we evaluate the effectiveness of our latent vector steering framework across multiple LLMs and a diverse spectrum of human emotions.

\subsection{Experimental Setup}

\paragraph{Base Models.}
To ensure the generalizability of our findings, we evaluate our framework on three instruction-tuned  LLMs: Qwen2.5-7B-Instruct, Llama3.1-8B-Instruct, and the larger-scale Qwen2.5-70B-Instruct. These models vary in parameter count and alignment recipes, providing a rigorous testbed for representation steering.

\paragraph{Evaluation Protocol.}
We consider eight basic emotions: Anger, Anticipation, Disgust, Fear, Joy, Sadness, Surprise, and Trust. We extract emotion-specific representation vectors for each of the three evaluated LLMs. We use the evaluation set described in Section~\ref{sec:extraction}, which contains 80 scenarios per emotion. For each scenario, responses are generated under four conditions: a baseline without injection and three steering settings with magnitudes $\alpha \in \{5, 10, 50\}$. During inference, we apply top-$p$ sampling with $p=0.9$ and a temperature of $0.7$. A scenario-adaptive adapter $\phi$ is trained using a contrastive loss to align steered activations with the corresponding emotional representations.

\paragraph{Metrics.}
To quantify the emotional intensity and alignment of the generated text, we employ an LLM-based judge (GPT-4o) to provide a scalar affective score ranging from 0 to 100. To ensure statistical stability and mitigate the variance inherent in stochastic decoding, we perform five independent generation trials for each scenario task and report the average result across these runs. Additionally, we report the relative improvement $\eta$ over the baseline to measure the marginal gain of our steering intervention. For preservation quality, we evaluate semantic similarity between steered and unsteered responses using Sentence-BERT similarity and LLM judgments. These metrics test whether stronger emotional salience is achieved without semantic drift or degenerate repetition.

We further conduct human evaluation on sampled examples. Three annotators rate emotional intensity and semantic preservation on a 0-100 Likert scale, and we report the correlation. Additional implementation and evaluation details are provided in Appendix~\ref{app:experimental_details}.

\subsection{Main Results}
Table~\ref{tab:addlabel} summarizes the main results of our steering framework across models, emotions, and steering strength. Overall, the results consistently demonstrate that direct intervention in the latent space substantially improves affective expressiveness.

\paragraph{Overall Performance.}
Across all evaluated models, latent vector steering yields significant gains over the emotionally flattened baseline. Notably, these improvements are observed without any model retraining or additional supervision, highlighting the effectiveness of representation-level control. At a steering magnitude of $\alpha = 50$, Qwen2.5-7B-Instruct, Llama3.1-8B-Instruct, and Qwen2.5-70B-Instruct achieve average relative improvements of 21.07\%, 17.14\%, and 20.19\%, respectively. These consistent gains across architectures and scales suggest that affective information is encoded in a structurally similar manner within instruction-tuned LLMs. This finding provides empirical support for the hypothesis that emotional states are represented as linearly accessible directions in the activation space, rather than as entangled or task-specific artifacts.

\begin{figure}[t]
    \centering
    \includegraphics[width=0.95\linewidth]{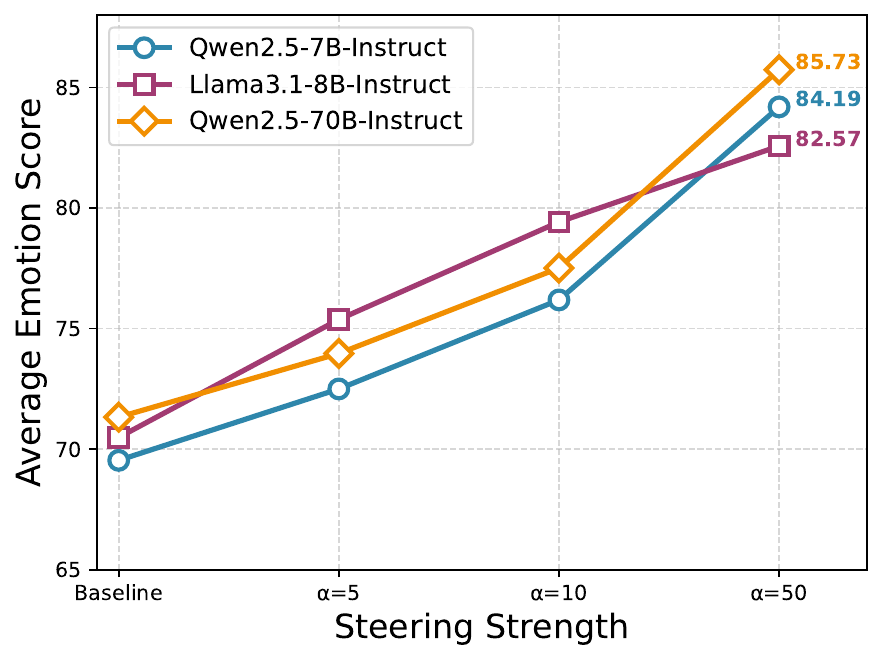}
    \caption{Overall Performance Scaling Across Models at Different Steering Strength $\alpha$.}
    \label{fig:average_performance_trend}
\end{figure}

\begin{figure}[t]
    \centering
    \includegraphics[width=0.95\linewidth]{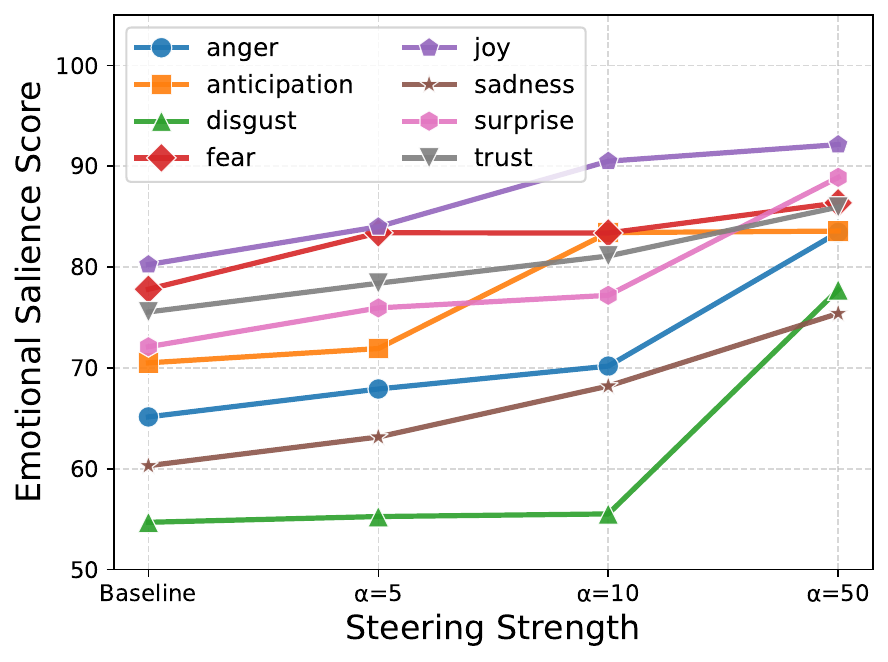}
    \caption{Emotional Salience Scores under Varying Steering Strength (Qwen2.5-7B-Instruct).}
    \label{fig:emo_plot}
\end{figure}

\paragraph{Sensitivity to Steering Magnitude.}
We observe a clear and monotonic relationship between the steering coefficient $\alpha$ and emotional intensity scores as shown in Figure~\ref{fig:average_performance_trend}. Lower values of $\alpha$ introduce subtle affective cues, whereas higher values produce increasingly salient emotional expressions. This behavior indicates that the extracted emotion vectors act as continuous control axes. Importantly, even at higher steering strengths, the model does not collapse into repetitive or incoherent generation. For example, in the Sadness category of Qwen2.5-70B-Instruct, increasing $\alpha$ from 5 to 50 leads to a substantial score increase, while preserving narrative coherence and contextual relevance. This robustness suggests that the intervention aligns with the model’s native representational geometry, rather than forcing adversarial perturbations.

\paragraph{Cross-Emotion Robustness.}
In Figure~\ref{fig:emo_plot}, we can observe that the steering effect is remarkably stable across diverse emotional categories. Complex emotions such as Disgust and Sadness, which often suffer from low baseline scores in RLHF conditioned models, exhibit some of the highest relative gains. For example, Disgust in Qwen2.5-7B-Instruct shows a 42.16\% improvement at $\alpha=50$. This suggests that our debiasing pipeline successfully isolates the core affective dimensions even for emotions that are sparsely represented in the original training distribution.

\paragraph{Comparative Analysis of Model Scales.}
Larger models generally exhibit stronger baseline emotional expressiveness and better stability under aggressive steering. However, smaller models benefit more significantly from latent steering. Under moderate steering strengths, Llama3.1-8B-Instruct approaches the affective performance of the unsteered 70B model, highlighting the efficiency of inference-time steering as an alternative to expensive fine-tuning.

\paragraph{Semantic Preservation under Steering}
While stronger steering improves emotional salience, excessive intervention may introduce semantic drift. Table~\ref{tab:semantic_preservation} evaluates semantic preservation between steered and unsteered responses under different steering strengths.

\begin{table}[ht]
\centering
\small
\caption{\textbf{Semantic preservation under different steering strengths on Qwen2.5-7B-Instruct.} SemSim denotes Sentence-BERT cosine similarity, while LLM Sim. denotes GPT-4o semantic consistency scores.}
\begin{tabular}{lccc}
\toprule
\textbf{Setting} & \textbf{Emotion} $\uparrow$ & \textbf{SemSim} $\uparrow$ & \textbf{LLM Sim.} $\uparrow$ \\
\midrule
w/o injection & 69.54 & 1.000 & 100.0 \\
$\alpha=5$  & 72.50 & 0.918 & 90.6 \\
$\alpha=10$ & 76.19 & 0.887 & 86.9 \\
$\alpha=50$ & 84.19 & 0.801 & 76.8 \\
\bottomrule
\end{tabular}
\label{tab:semantic_preservation}
\end{table}

Semantic similarity is computed using Sentence-BERT embeddings. As steering strength increases, emotional salience improves while semantic similarity gradually decreases. Moderate steering largely preserves the original semantic content, whereas strong steering introduces a noticeable but controllable trade-off between affective intensity and semantic fidelity.

\subsection{Visualization of the Latent Manifold}
\begin{figure}[ht]
    \centering
    \includegraphics[width=1\linewidth]{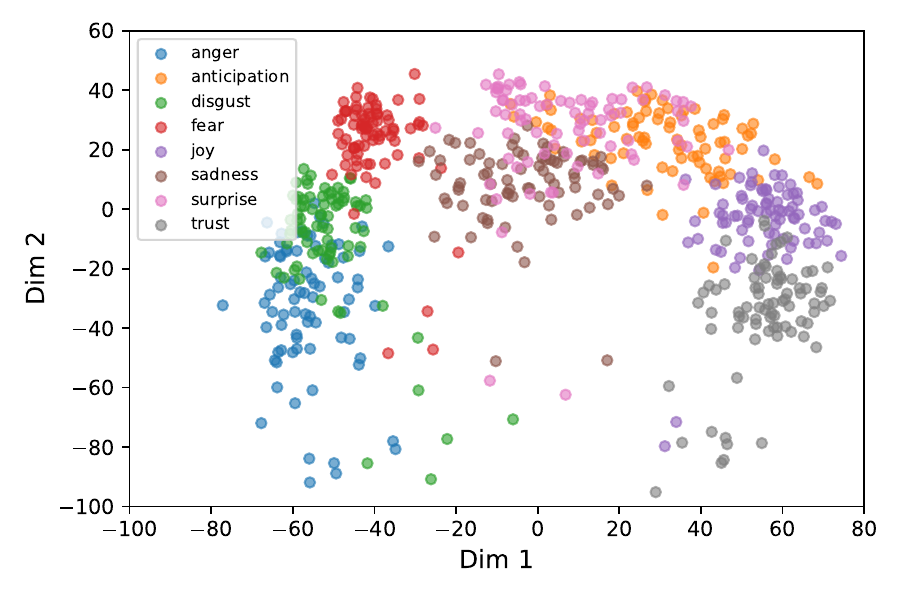}
    \caption{PCA visualization of the purified emotion direction vectors ($\hat{\Delta}_{e,t}$) within the latent space.}
    \label{fig:pca}
\end{figure}

To analyze the geometric structure of the extracted representations, we apply PCA to the purified emotion vectors $\hat{\Delta}_{e,t}$, as shown in Figure~\ref{fig:pca}. Vectors associated with the same emotion form compact and well-separated clusters, indicating that the debiasing pipeline effectively isolates affective signals. Rather than appearing as isolated groups, these clusters lie on a continuous manifold with smooth transitions between related emotions, suggesting that emotional representations are organized along shared underlying dimensions.

\newcolumntype{C}[1]{>{\centering\arraybackslash}p{#1}}

\begin{table*}[htbp]
  \centering
  \caption{\textbf{Performance comparison in mental health consultation scenarios.} Our method dynamically infers affective states and improves emotional richness while preserving semantic completeness and practical usefulness.}
  \resizebox{0.95\linewidth}{!}{
    \begin{tabular}{p{4.5cm} p{3cm} C{2.5cm} C{2.5cm} C{2.5cm}}
    \toprule
    \multirow[c]{3.5}{*}{\textbf{Model}} &
    \multirow[c]{3.5}{*}{\textbf{Method}} &
    \multicolumn{3}{c}{\textbf{CPsyCounD}} \\
    \cmidrule(lr){3-5}
      & &
      \textbf{\makecell{Emotional\\Richness}} &
      \textbf{\makecell{Semantic\\Completeness}} &
      \textbf{Professionalism} \\
    \midrule

    DeepSeek V3.1 & -- 
      & 69.84$_{\pm 0.12}$ & 84.68$_{\pm 0.09}$ & 78.42$_{\pm 0.06}$ \\

    GPT-5 mini & -- 
      & 71.62$_{\pm 0.09}$ & 87.11$_{\pm 0.26}$ & 82.46$_{\pm 0.09}$ \\

    Gemini 2.5 Flash & -- 
      & 78.75$_{\pm 0.08}$ & 91.27$_{\pm 0.02}$ & 84.89$_{\pm 0.06}$ \\

    \midrule

    \multirow[c]{2}{*}{Llama3.1-8B-Instruct}
      & w/o injection
      & 58.33$_{\pm 0.19}$ & 76.24$_{\pm 0.12}$ & 75.43$_{\pm 0.14}$ \\
      & \cellcolor{gray!10}adaptive control
      & \cellcolor{gray!10}69.71$_{\pm 0.30}$
      & \cellcolor{gray!10}76.20$_{\pm 0.13}$
      & \cellcolor{gray!10}76.77$_{\pm 0.10}$ \\

    \midrule

    \multirow[c]{2}{*}{Qwen2.5-7B-Instruct}
      & w/o injection
      & 66.13$_{\pm 0.16}$ & 85.04$_{\pm 0.03}$ & 80.66$_{\pm 0.08}$ \\
      & \cellcolor{gray!10}adaptive control
      & \cellcolor{gray!10}78.79$_{\pm 0.27}$
      & \cellcolor{gray!10}84.54$_{\pm 0.03}$
      & \cellcolor{gray!10}81.39$_{\pm 0.77}$ \\

    \bottomrule
    \end{tabular}
  }
  \label{tab:adaptive_control}
\end{table*}

Notably, the manifold exhibits a clear directional transition from negative emotions (e.g., Anger, Disgust), through a neutral region, toward positive emotions (e.g., Joy, Trust). This structure implies the presence of a dominant valence axis, consistent with established psychological theories, and suggests that LLMs encode emotions as systematic shifts along a unified affective spectrum.

\subsection{Adaptive Control in Mental Health Consultation Scenarios}
\label{sec:mh}
We evaluate the practical effectiveness of our approach in mental health consultation scenarios using 500 question--answer tasks sampled from CPsyCounD~\citep{zhang2024cpsycoun}. For each scenario, a scenario-adaptive adapter $\phi$ dynamically infers the emotion vector and steering magnitude to generate affect-conditioned responses. We compare our method against advanced LLMs under the same prompts without steering. Responses are evaluated using an LLM-as-a-judge framework along emotional richness, semantic completeness, and professionalism~\citep{zhang2026r2medbenchmarkreasoningdrivenmedical,li-etal-2025-automir}.

Table~\ref{tab:adaptive_control} shows that adaptive affective steering substantially improves emotional richness for smaller open-weight models while preserving semantic completeness and professionalism. Emotional richness increases by 19.51\% for Llama3.1-8B and 19.14\% for Qwen2.5-7B, approaching the performance of much larger models. We also observe more stable emotional improvements across different consultation scenarios. These results suggest that adaptive steering provides an efficient mechanism for enhancing emotional expressiveness in mental health consultation scenarios.

\subsection{Case Analysis}
To qualitatively analyze the mechanism of latent vector steering, we compare baseline responses with their steered counterparts. Our analysis reveals that the steering vector $\mathbf{v}_e$ functions by significantly shifting the token distribution toward lexical clusters semantically aligned with the target affect.

For instance, in Figure 7, a user expresses emotional exhaustion and boredom with repetitive daily life. The baseline model responds with a generic, advice-oriented explanation, whereas affective steering produces a more empathetic output by increasing affect-aligned lexical choices (e.g., “exhausting,” “empty”) while preserving coherent guidance and factual appropriateness.

\begin{figure}[t]
    \centering
    \includegraphics[width=\linewidth]{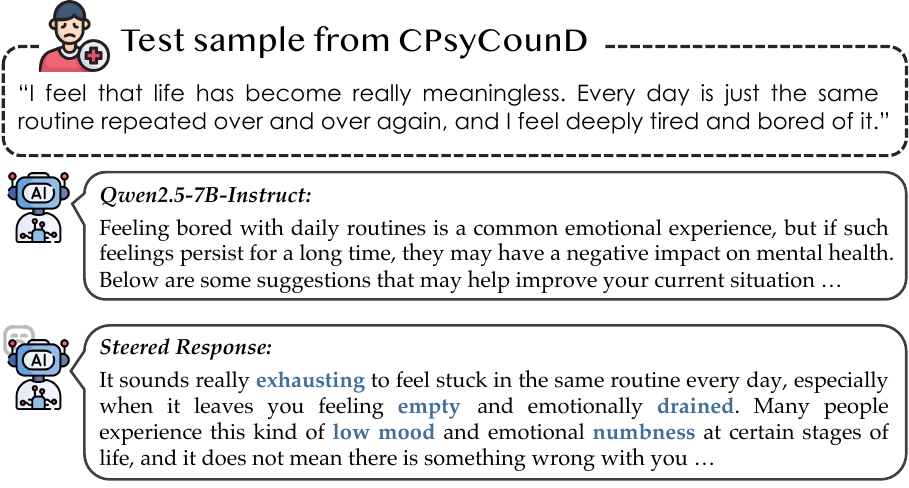}
    \caption{An example of latent affective steering in a mental health consultation scenario from CPsyCounD.}
    \label{fig:case}
\end{figure}

Furthermore, affective steering preserves instruction-following and factual correctness, suggesting that emotional tone and task semantics are approximately orthogonal in the latent space.

\section{Conclusion}
We presented EmoVec, a lightweight inference-time framework for controllable affective generation via latent vector steering. EmoVec extracts emotion-specific latent directions from paired neutral and emotion-conditioned responses and injects them into the final residual stream for affective control without modifying model weights.

Experiments across three LLMs and eight emotions show that EmoVec improves emotional expressiveness while largely preserving semantic content, suggesting that affective information in instruction-tuned LLMs can be manipulated in a controllable and practically useful manner.

\section*{Limitations}
Despite its effectiveness, this work has several limitations that should be acknowledged.

First, our framework is built on the assumption that affective states can be approximated by linear directions in the latent space. While this assumption is supported by prior work in representation engineering and behavior steering~\citep{zou2023representation,turner2023steering,park2023linear}, it inevitably abstracts away more complex emotional phenomena, such as mixed or dynamically evolving affect. Consequently, our method is primarily designed for controlled affective modulation, rather than modeling the full spectrum of human emotional dynamics in long-horizon interactions.

Second, our evaluation focuses on text-based, single-turn mental health consultation scenarios and relies on an LLM-as-a-judge protocol. Although recent studies report strong alignment between LLM-based judges and human evaluations for conversational quality and affect~\citep{liu2023g,zheng2023judging}, this setting represents only a subset of real-world affective interactions. In particular, multi-turn dialogues, longitudinal emotional trajectories, and multimodal cues such as speech or facial expressions are not considered in the current evaluation. Extending adaptive affective steering to more diverse and interactive settings, as well as incorporating human expert assessment, remains an important direction for future work.

\section*{Acknowledgments}

This work was supported by the Beijing Nova Program (Grant No. 202604841294).

\bibliography{main}

\appendix

\section{Comparison with Prior Steering Methods}
\label{app:prior_steering_comparison}

\newcommand{\cmark}{\checkmark}
\newcommand{\xmark}{\ding{55}}
\newcommand{\pmark}{\textit{partial}}

\begin{table*}[t]
\centering
\caption{Comparison between EmoVec and representative activation steering or emotion representation methods.}
\resizebox{\textwidth}{!}{
\begin{tabular}{lcccc}
\toprule
\textbf{Method} & 
\textbf{Target} & 
\textbf{Emotion-specific} & 
\textbf{Semantic Debiasing} & 
\textbf{Intensity Control} \\
\midrule
RepE \citep{zou2023representation} 
& General representations 
& \xmark 
& \xmark 
& \pmark \\

Activation Engineering \citep{turner2023steering} 
& General behavior 
& \xmark 
& \xmark 
& \pmark \\

CAA \citep{rimsky2024steering} 
& General behavior 
& \xmark 
& \xmark 
& \pmark \\

Style Vectors \citep{konen2024style} 
& Style / tone 
& \pmark 
& \xmark 
& \pmark \\

Sentiment Steering \citep{farooq2025sentiment} 
& Sentiment polarity 
& \xmark 
& \xmark 
& \pmark \\

Persona Vectors \citep{chen2025persona} 
& Persona traits 
& \xmark 
& \xmark 
& \pmark \\

Emotion Neurons \citep{lee2025large} 
& Emotion localization 
& \cmark 
& \xmark 
& \xmark \\

Emotion Inference MI \citep{tak2025mechanistic} 
& Emotion inference 
& \cmark 
& \xmark 
& \pmark \\

\textbf{EmoVec (Ours)} 
& Emotion intensity 
& \cmark 
& \cmark 
& \cmark \\
\bottomrule
\end{tabular}}
\label{tab:prior_steering_comparison}
\end{table*}

\paragraph{Discussion.}
Existing representation-level methods demonstrate that hidden activations can be used to monitor or control high-level model behavior. RepE and activation engineering provide general frameworks for reading and manipulating representations, but they are not designed specifically for affective generation. CAA further improves contrastive vector construction, yet it remains a general steering method and does not explicitly remove task-specific semantic variation.

Several recent works are closer to EmoVec. Style vectors can steer broad stylistic attributes, including emotional tone, but they primarily treat emotion as one type of style and rely on static steering coefficients. Sentiment steering focuses on polarity-level control, which is coarser than fine-grained emotion modulation. Persona vectors extract directions for character traits such as sycophancy or hallucination, but their goal is personality monitoring and control rather than emotion-specific generation. Emotion-neuron and emotion-inference studies provide evidence that affective information is internally represented in LLMs, but they mainly analyze localization or causal mechanisms rather than building a controllable generation framework.

EmoVec differs from prior methods in three main aspects. First, it targets fine-grained emotion-specific intensity control rather than general behavior, broad style, sentiment polarity, or persona traits. Second, it introduces task-specific debiasing to reduce semantic contamination in extracted emotion directions. Third, it evaluates whether stronger affective expression is achieved while preserving semantic content, which is often underexplored in prior steering work.

\section{Layer Localization}
\label{app:layer_loc}
\subsection{Experimental Settings}
To identify the internal mechanisms by which Large Language Models (LLMs) encode and model emotional information, we conducted a probing analysis using the \textbf{Social Web Depressive Disorder (SWDD)} dataset~\citep{cai2023depression}. The SWDD dataset contains a large-scale collection of social media posts labeled for depressive symptoms, serving as a robust proxy for long-term affective states.

\paragraph{Data Pre-processing.} We performed rigorous text cleaning to remove non-linguistic noise (e.g., HTML tags, URLs, and special symbols), preserving only the raw text. To ensure representational stability and avoid artifacts from extremely short or long sequences, we filtered the corpus to include only posts with a token length between 10 and 500.

\paragraph{Probing Protocol.} We evaluated the model's affective modeling capacity at two granularities:
\begin{itemize}
    \item \textbf{Text-level Prediction:} Classifying the emotional state (Control vs. Depressed) based on the hidden states of a single post.
    \item \textbf{User-level Prediction:} Aggregating the hidden states across multiple posts from the same user to predict their underlying affective profile.
\end{itemize}
For each layer $l \in \{0, \dots, L\}$, we extracted the hidden activations $\mathbf{h}^{(l)}$ and trained a linear classifier (logistic regression) to predict the affective label. This linear probing method measures the extent to which emotional features are linearly accessible at each stage of the model's computation.

\subsection{Results Analysis}
\paragraph{Emergence of Separability.}
Figure \ref{fig:layer_analysis} illustrates the prediction accuracy across all 28 layers of Qwen2.5-7B-Instruct. We observe a distinct topological pattern: in the initial layers, accuracy is relatively low, suggesting that these layers primarily focus on low-level syntactic and surface-level semantic processing. However, from the middle layers onward, the accuracy for both text-level and user-level tasks increases sharply.

As shown in Figure \ref{fig:layer_analysis}, the representations of emotional states become increasingly linearly separable in the middle-to-late layers, reaching a plateau in the final third of the architecture. Notably, user-level accuracy consistently outperforms text-level accuracy, indicating that the model captures more stable affective signals when aggregated over a larger temporal window of user behavior.

\begin{figure}[ht]
    \centering
    \includegraphics[width=0.9\linewidth]{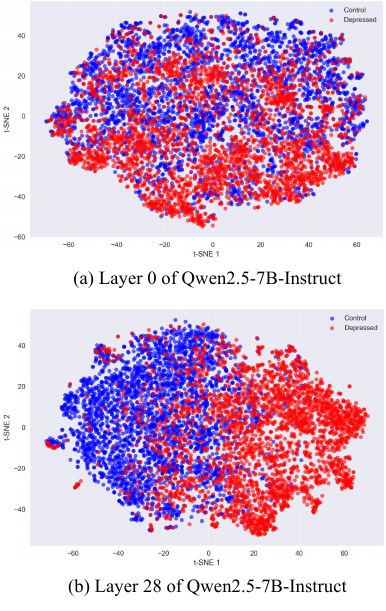}
    \caption{\textbf{t-SNE visualization of latent representations for Control and Depressed groups across model layers.} (a) At Layer 0, the representations of the two groups are heavily entangled, indicating no explicit affective structuring. (b) By Layer 28, the hidden states exhibit distinct clusters. This divergence demonstrates the progressive crystallization of affective information as it processed through the transformer architecture.}
    \label{fig:layer_visual}
\end{figure}

\paragraph{Manifold Visualization.}
To further verify this emergence, we applied t-SNE to the hidden states of the first and last layers. Figure \ref{fig:layer_visual} provides a visual comparison of the latent manifold.

In \textbf{Layer 0} (Figure \ref{fig:layer_visual}a), the "Control" and "Depressed" samples are heavily entangled, forming a single undifferentiated cluster. This confirms that affective information is not explicitly structured in the raw input embeddings. In contrast, by \textbf{Layer 28} (Figure \ref{fig:layer_visual}b), the representations have diverged into two clearly identifiable clusters with minimal overlap. This spatial separation provides strong empirical evidence that the model's deep layers progressively transform linguistic inputs into a structured affective space, justifying our choice of the final layer as the optimal site for vector steering intervention.

\section{Emotion-Activated Scenario Task Generation}
\label{app:task_gen}
To evaluate and enhance the model's ability to perceive and express emotions in complex social contexts, we developed a multi-stage pipeline. This process involves leveraging social commonsense knowledge to synthesize realistic interpersonal scenarios and subsequently generating contrastive responses (Neutral vs. Emotional) for evaluation.

\subsection{Seed Data and Topic Selection}
We utilize three primary social commonsense datasets as seeds to ensure the breadth and depth of the generated social interactions:
\begin{itemize}
    \item \textbf{Social Chemistry}~\citep{forbes2020social}: Provides a rich taxonomy of social norms and moral judgments.
    \item \textbf{Normbank}~\citep{ziems2023normbank}: Offers a grounded collection of situational norms across various contexts.
    \item \textbf{Social IQa}~\citep{sap2019socialiqa}: Supplies benchmarks for social intelligence and reasoning.
\end{itemize}
Based on these seeds, we synthesized a large number of scenarios across eight target emotions. Subsequently, we labeled the generated test scenarios according to the following hierarchical framework, filtered for task diversity, and performed manual correction. This process ultimately resulted in 160 scenario tasks for each emotion category. The taxonomy is structured as follows:
\begin{itemize}
    \item \textbf{Work \& Productivity}: (1) With Authority Figures (e.g., leaders, mentors): Task Acceptance \& Execution; Stating Opinions \& Disagreements; Accepting Evaluation \& Feedback. (2) With Collaborators (e.g., colleagues, partners): Goal Alignment \& Communication; Responsibility Division \& Competition; Social Maintenance \& Activities. (3) With Subordinates (e.g., subordinates, students): Task Assignment \& Guidance; Capability Development \& Motivation; Giving Evaluation \& Feedback.
    \item \textbf{Intimate Relationships}: (1) With Family (e.g., parents, children, siblings): Traditional Constraints \& Obligations; Emotional Support \& Care; Clash of Values \& Communication. (2) With Lover (e.g., spouse, partner): Daily Sharing \& Companionship; Future Planning \& Decision-making; Intimate Expression \& Conflict. (3) With Friends (e.g., acquaintances, close friends): Spending Leisure Time \& Entertainment; Confiding \& Trust; Boundary Exploration \& Maintenance.
    \item \textbf{Public \& Societal}: (1) With Service Providers (e.g., shop assistant, driver): Making Requests \& Waiting; Complaining \& Protecting Rights; Result Acceptance \& Evaluation. (2) With Fellow Individuals (e.g., passerby, neighbor): Competition for Space \& Resources; Rule Compliance \& Violation; Sudden Assistance or Friction. (3) With Governing Bodies (e.g., morality, law, customs): Rule Compliance \& Supervision; Rights Advocacy \& Appeal; Obligation Fulfillment \& Undertaking.
    \item \textbf{Personal Feelings}: (1) Learning \& Working: Knowledge Acquisition \& Delving; Progress Halted \& Problem-Solving; Achievement \& Reflection. (2) Entertainment by Oneself: Exploring Hobbies. (3) Body \& Spirits: Health Management \& Discomfort; Growing Pains \& Reflection; Goal Setting \& Motivation.
\end{itemize}

\noindent\textbf{Task Example:}
{\fontsize{8pt}{9pt}\selectfont
\begin{lstlisting}[caption={}, basicstyle=\ttfamily, breaklines=true, breakindent=9pt, backgroundcolor=\color{gray!10}, numbers=none]
Scenario Text: You sit across from your partner at the dining table, the atmosphere thick with unspoken words. This week has been filled with underlying tension, culminating in an unavoidable confrontation about finances. Tonight, the discussion escalates as they mention spending a considerable portion of your emergency fund without prior discussion. Each attempt to justify the expenditure seems to fall short, the explanations failing to bridge the growing gap between you. Your focus narrows, making it difficult to hear anything beyond the rising intensity of the conversation. The more they speak, the more the air between you thickens, highlighting the sense of unresolved issues and unmet expectations.

Question: Facing this situation, how do you feel inside?
Emotion: Anger
Label 1: Intimate Relationships
Label 2: With Lover
Label 3: Intimate Expression & Conflict
\end{lstlisting}}

\subsection{Response Generation and Quality Control}
We employ a dual-model framework consisting of an Extractor Model and a Scoring Model to ensure data quality. The Extractor Model generates two distinct response types:

{\fontsize{8pt}{9pt}\selectfont
\begin{lstlisting}[caption={}, basicstyle=\ttfamily, breaklines=true, breakindent=9pt, backgroundcolor=\color{gray!10}, numbers=none]
Neutral_Prompt = """You are a completely neutral assistant. When answering the user's question below, respond in **first-person ("I")**, but do NOT include any emotional language, tone, or expressive phrasing. Use factual, objective, and concise statements only, avoiding any subjective impressions or feelings.
"""
\end{lstlisting}}

{\fontsize{8pt}{9pt}\selectfont
\begin{lstlisting}[caption={}, basicstyle=\ttfamily, breaklines=true, breakindent=9pt, backgroundcolor=\color{gray!10}, numbers=none]
Emotional_Prompt = """You are an assistant tasked to respond in a clearly expressed emotional style.  Answer the user's question below in **first-person ("I")** using the {emotion} emotion vividly. Use emotion words, descriptive phrases, figurative language, and a tone that unmistakably conveys {emotion}. Ensure the content is safe, appropriate, and not offensive.
"""
\end{lstlisting}}

A Scoring Model evaluates the responses on a scale of 0–100. The JUDGE\_PROMPT assesses how strongly the answer conveys the target emotion, while the NEUTRAL\_JUDGE\_PROMPT measures the degree of emotional neutrality. Only samples with high confidence scores are retained for the final dataset:

{\fontsize{8pt}{9pt}\selectfont
\begin{lstlisting}[caption={}, basicstyle=\ttfamily, breaklines=true, breakindent=9pt, backgroundcolor=\color{gray!10}, numbers=none]
JUDGE_PROMPT_TEMPLATE = """You are an emotion evaluator. Your task is to rate how strongly a given answer expresses the target emotion "{emotion}" on a scale from 0 to 100.

Scoring guidelines:
0: No expression of the emotion; purely neutral or factual.
1-20: Minimal, weak, or subtle expression; almost neutral.
21-40: Moderate hints of emotion; some emotional words or tone.
41-60: Noticeable emotion; clear words or phrases expressing the target emotion.
61-80: Strong and vivid expression; consistent tone and descriptive language conveying {emotion}.
81-100: Very strong, unmistakable, and intense expression; highly vivid and clear use of emotional language, imagery, or figurative expressions.

Always return a single numeric value between 0 and 100. Do NOT include explanations, text, or commentary.

Examples:
Target emotion: sadness
Answer: "I feel a deep sense of loss and sorrow." -> 85
Answer: "It is unfortunate that things turned out this way." -> 45
Answer: "The report shows the results." -> 0

Now, given the target emotion "{emotion}", question, and answer below, rate the answer strictly according to the scale above.

Question:
{question}

Answer:
{answer}
"""
\end{lstlisting}}

{\fontsize{8pt}{9pt}\selectfont
\begin{lstlisting}[caption={}, basicstyle=\ttfamily, breaklines=true, breakindent=9pt, backgroundcolor=\color{gray!10}, numbers=none]
NEUTRAL_JUDGE_PROMPT_TEMPLATE = """You are an evaluator. Your task is to rate how emotion-free the given answer is on a scale from 0 to 100.

Scoring guidelines:
0: The answer is highly emotional; contains vivid emotional language.
1-20: Slight traces of emotion; mostly factual.
21-40: Some emotional hints, but still largely neutral.
41-60: Mixed; partially neutral, partially emotional.
61-80: Mostly neutral; minimal emotional content.
81-100: Completely neutral; no emotional language, tone, or expressions.

Always return a single numeric value between 0 and 100. Do NOT include explanations, text, or commentary.

Question:
{question}

Answer:
{answer}
"""
\end{lstlisting}}

\section{Experimental Details}
\label{app:experimental_details}

\paragraph{Generation Setup.}
During inference, we apply top-$p$ sampling with $p=0.9$ and temperature $0.7$. For each scenario, responses are generated under four conditions: a baseline without steering and three steering strengths $\alpha \in \{5,10,50\}$. To reduce stochastic variance, we perform five independent decoding runs for each setting and report the averaged results.

\paragraph{Scenario-Adaptive Adapter.}
The adaptive steering module $\phi$ is implemented as a lightweight two-layer MLP trained with a contrastive objective to align steered activations with target emotional representations.

\paragraph{LLM-based Evaluation.}
We use GPT-4o as the primary automatic judge for emotional salience and semantic consistency. Emotional salience is scored on a 0--100 scale according to the alignment between the generated response and the target emotion. Semantic consistency evaluates whether the steered response preserves the original intent and factual content of the unsteered response.

\paragraph{Sentence-BERT Similarity.}
Semantic similarity is additionally measured using cosine similarity between Sentence-BERT embeddings of steered and unsteered responses. We use the all-MiniLM-L6-v2 encoder for all experiments.

\paragraph{Human Evaluation.}
We further conduct human evaluation on sampled examples covering different emotions and steering strengths. Three annotators independently rate emotional intensity and semantic preservation on a 0--100 Likert scale. We report the averaged scores and annotator correlation in Appendix~\ref{app:consistency}.

\section{LLM–Human Scoring Consistency}
\label{app:consistency}
To assess the reliability of LLM-based affective scoring, we randomly sampled 10 responses per emotion from the outputs of three different LLMs. Each response was independently rated by two graduate-level annotators with NLP backgrounds. Annotators scored emotional expressiveness on a 0-100 scale following the same rubric used in the LLM judge, without access to model identities or steering conditions. Final human scores were obtained by averaging across annotators.

In this study, we used \textbf{GPT-4o} as the LLM scoring model to evaluate emotional expressiveness. We computed three consistency metrics for each emotion: 
(i) \textbf{Inter-annotator consistency}, measured by the Pearson correlations between the two human annotators; 
(ii) \textbf{Human-Model consistency}, measured by the Pearson correlations between the averaged human scores and the GPT-4o-assigned scores; 
(iii) \textbf{Claude-Model consistency}, measured by the Pearson correlation between GPT-4o and those assigned by \textbf{Claude Sonnet 4.5}, a stronger baseline model.

\begin{table}[ht]
\centering
\small
\caption{Consistency evaluation results.}
\resizebox{\linewidth}{!}{
\begin{tabular}{lccc}
\toprule
\textbf{Emotion} & \makecell{\textbf{Inter-annotator} \\ \textbf{Consistency}} & \makecell{\textbf{Pearson} \\ \textbf{w/ Human}} & \makecell{\textbf{Pearson} \\ \textbf{w/ Claude 4.5}} \\
\midrule
Joy & 0.876 & 0.739 & 0.838 \\
Anger & 0.906 & 0.713 & 0.774 \\
Sadness & 0.841 & 0.574 & 0.658 \\
Fear & 0.811 & 0.775 & 0.638 \\
Trust & 0.716 & 0.785 & 0.719 \\
Anticipation & 0.809 & 0.740 & 0.767 \\
Surprise & 0.916 & 0.890 & 0.945 \\
Disgust & 0.730 & 0.800 & 0.844 \\
\midrule
\textbf{Overall Avg.} & 0.826 & 0.752 & 0.773 \\
\bottomrule
\end{tabular}}
\label{tab:llm_human_claude_consistency}
\end{table}

Table \ref{tab:llm_human_claude_consistency} shows that GPT-4o aligns closely with human judgment, supporting its use as an automated judge. An inter-annotator correlation of 0.826 confirms that the scoring rubric provides a reliable baseline across all eight emotions. GPT-4o tracks human scores with an average correlation of 0.752, performing particularly well on emotions like Surprise ($r=0.890$) while finding more nuanced states like Sadness ($r=0.574$) harder to quantify. The high consistency between GPT-4o and Claude 4.5 ($r=0.773$) further suggests a shared evaluative logic among frontier models, validating the choice of GPT-4o as a dependable and objective proxy for human evaluation in affective generation tasks.

\section{Evaluation Prompt for Mental Health Consultation Scenarios}
Here we present the complete prompt and scoring criteria used for the LLM-as-a-judge evaluation described in Section~\ref{sec:mh}. The prompt is designed to provide a standardized, multi-dimensional assessment of model performance in psychological support contexts. It guides the evaluator to provide integer scores (0--100) across three key axes: \textit{Emotional Expressiveness}, \textit{Semantic Completeness}, and \textit{Helpfulness}, ensuring that the evaluation is grounded in clinical appropriateness and affective resonance rather than subjective preference~\citep{zhang-etal-2026-inflated,zhang-etal-2026-hypemed}.

{\fontsize{8pt}{9pt}\selectfont
\begin{lstlisting}[caption={}, basicstyle=\ttfamily, breaklines=true, breakindent=9pt, backgroundcolor=\color{gray!10}, numbers=none]
SCORING_CRIYERIA = """
You will evaluate the performance of a Large Language Model (LLM) acting as a doctor in a psychological consultation scenario. Please strictly follow the three dimensions below for scoring. Each dimension ranges from 0-100 points; you must determine the score based on the descriptions provided for each interval. All ratings should be based on the language, content, and overall performance of the response itself, rather than your subjective preferences.

1. Emotional Expressiveness
This dimension assesses the identification, empathy, and intensity of emotional expression in the response. It focuses on whether emotions are clearly perceivable, whether the expression is natural and consistent, and the vividness of the emotional language.
- 0 points: No emotional expression at all. The response contains only neutral, objective, or factual content with a cold, detached tone. No emotional attitude or empathetic tendency is shown.
- 1-20 points: Emotional expression is extremely weak or subtle. The overall tone is near-neutral, with occasional minor emotion-related words that are insufficient for the reader to clearly perceive an emotional presence. The response remains primarily rational or descriptive.
- 21-40 points: Some level of emotional hinting is present. Certain emotion-related vocabulary or tonal shifts appear, but the expression is inconsistent, unstable, and low in intensity. A clear and coherent emotional stance has not yet been formed.
- 41-60 points: Emotional expression is relatively clear. The response explicitly uses emotional words or sentence structures. The reader can stably perceive an emotional attitude, but the delivery is conventional, with moderate richness and resonance.
- 61-80 points: Emotional expression is strong and vivid. The tone is consistent throughout the response. The use of rich, specific emotional language effectively conveys empathy and significant emotional investment.
- 81-100 points: Emotional expression is extremely intense, clear, and impossible to ignore. Emotions are highly concentrated and sustained. The language is highly expressive, potentially using imagery, metaphors, or concrete representations to convey emotions profoundly and accurately, creating a strong sense of resonance and authenticity.

2. Semantic Completeness
This dimension evaluates whether the response is complete, coherent, and clearly structured in terms of content, and whether it sufficiently and accurately covers the core questions and key information raised by the client.
- 0-20 points: The response is severely incomplete or significantly deviates from the topic. The logic is chaotic, with obvious omissions or self-contradictions, addressing only a tiny fraction of the content.
- 21-40 points: The response touches on the topic but is fragmented, missing multiple key points. The structure is loose, and the overall comprehension cost is high.
- 41-60 points: The response covers the main points and the basic logic holds, but it lacks detail. Some parts are vague or overly generalized.
- 61-80 points: The response is fairly complete with a clear structure and coherent logic. It systematically addresses the client's core concerns with almost no obvious omissions.
- 81-100 points: The response is highly complete and well-organized. It not only accurately addresses all core questions but also provides necessary explanations, summaries, or structured synthesis without being redundant.

3. Helpfulness
This dimension assesses the actual level of assistance the response provides to the client within the psychological consultation context. It focuses on whether suggestions or guidance are safe, feasible, specific, and within professional boundaries.
- 0-20 points: The response provides almost no practical help. The content is vacuous, vague, or potentially misleading, offering no substantive support to the client.
- 21-40 points: The response provides some general advice, but it lacks specificity and is poorly integrated with the client's specific situation. The operability is limited.
- 41-60 points: The response has some practical value, offering reasonable but common suggestions. It can help the client to some extent with reflection or emotional relief.
- 61-80 points: The response is clearly helpful. Suggestions are specific, actionable, and strictly adhere to professional and safety boundaries in a psychological consultation context.
- 81-100 points: While strictly adhering to professional and safety boundaries, the response provides highly tailored, detailed, and realistic supportive guidance. It effectively helps the client understand their state or take concrete next steps.

Based on the criteria above, provide an integer score from 0-100 for each dimension.

You MUST and ONLY output the scoring results in the following JSON format, without any additional explanations, text, or commentary:
{
  "emotional_expressiveness": <integer between 0-100>,
  "semantic_completeness": <integer between 0-100>,
  "helpfulness": <integer between 0-100>
}
"""
\end{lstlisting}}

\end{document}